# Genotype-Guided Radiomics Signatures for Recurrence Prediction of Non-Small-Cell Lung Cancer

Panyanat Aonpong, Yutaro Iwamoto, Xian-Hua Han, Lanfen Lin*, and Yen-Wei Chen*, *Member, IEEE*

**Abstract**— Non-small cell lung cancer (NSCLC) is a serious disease and has a high recurrence rate after the surgery. Recently, many machine learning methods have been proposed for recurrence prediction. The methods using gene data have high prediction accuracy but require high cost. Although the radiomics signatures using only CT image are not expensive, its accuracy is relatively low. In this paper, we propose a genotype-guided radiomics method (GGR) for obtaining high prediction accuracy with low cost. We used a public radiogenomics dataset of NSCLC, which includes CT images and gene data. The proposed method is a two-step method, which consists of two models. The first model is a gene estimation model, which is used to estimate the gene expression from radiomics features and deep features extracted from computer tomography (CT) image. The second model is used to predict the recurrence using the estimated gene expression data. The proposed GGR method designed based on hybrid features which is combination of handcrafted-based and deep learning-based. The experiments demonstrated that the prediction accuracy can be improved significantly from 78.61% (existing radiomics method) and 79.14% (deep learning method) to 83.28% by the proposed GGR.

─────────────── ◆ ───────────────

## 1 INTRODUCTION

Non-small cell lung cancer (NSCLC) is one of the most dangerous diseases in humans [1]. The NSCLC occurs about 80% - 85% of lung cancer patients [1] and can be treated by surgery [1, 2]. However, after the surgery, the NSCLC has a high recurrence rate [2, 3]. From the statistics, about 50% of patients die from tumor recurrence [4]. Due to this, preoperative recurrence prediction is helpful. If doctors and patients have accurate prediction of the risk of recurrence, the doctor will be able to prepare and look after patients appropriately [5].

Several machine learning methods have recently been proposed or developed for preoperative prediction of NSCLC recurrences based on genomics and radiomics information (Computed Tomography images; CT-images). Existing methods can be categorized into three types: (1) gene expression-based methods [4, 6], (2) radiomics signatures (using CT-image)-based methods [7-10, 12-14] and (3) fused gene expression and radiomics signatures-based methods [3, 10]. The first and third types use gene expression information for prediction. These methods have high prediction accuracy. In particular, the third type, which uses the combination of genomics and radiomics, has produced the highest results. The limitations of these methods are their high cost due to the complexity of the examination [11] and they are invasive diagnostic methods. In many cases, gene expression data is not available. On the other hand, since the second type of radiomics-based methods use only radiomics information (CT images), these methods are non-invasive diagnostic methods and do not cost much compared to gene testing. They can be used in most cases. The limitation of these methods is a low prediction accuracy. Recently, many researches tried to improve the accuracy of the radiomics-based methods by using more complicate deep models with convolutional products instead of radiomics signatures [3, 4, 6-9, 12-14], for example the use of ResNet50 [13, 14], DenseNet121 [15, 16] and also the fusion of deep features and handcrafted features [12, 16].

In this research, we propose a genotype-guided radiomics-based approach to improve the prediction accuracy of radiomics-based methods. The proposed method consists of two models. The first model is gene estimation. This model is used to estimate the gene expression information from CT images. The second model, recurrence prediction, is used to predict recurrence based on estimated gene expressions. We trained a model that maps the CT images and gene expressions by using pairs of CT images and gene expressions as the training data. By using the mapping function, we can predict gene expression from CT images (test data). Then the predicted gene expressions are used

- *Panyanat Aonpong is with the College of Information Science and Engineering, Ritsumeikan University, Shiga, Japan. E-mail: gr0399rh@ritsumei.ac.jp*
- *Yutaro Iwamoto is with the College of Information Science and Engineering, Ritsumeikan University, Shiga, Japan. E-mail: yiwamoto@fc.ritsumei.ac.jp*
- *Xian-Hua Han is with Artificial Intelligence Research Center, Yamaguchi University, Japan. E-mail: hanxhua@yamaguchi-u.ac.jp*
- *Lanfen Lin is with the College of Computer Science and Technology, Zhejiang University, Hangzhou, China. Email: llf@zju.edu.cn*
- *Yen-Wei Chen is with the College of Information Science and Engineering, Ritsumeikan University, Shiga, Japan, Zhejiang Lab, Hangzhou, China and College of Computer Science and Technology, Zhejiang University, Hangzhou, China. E-mail: chen@is.ritsumei.ac.jp*
- *\*Corresponding Authors: Yen-Wei Chen (chen@is.ritsumei.ac.jp), Lanfen Lin (llf@zju.edu.cn)*

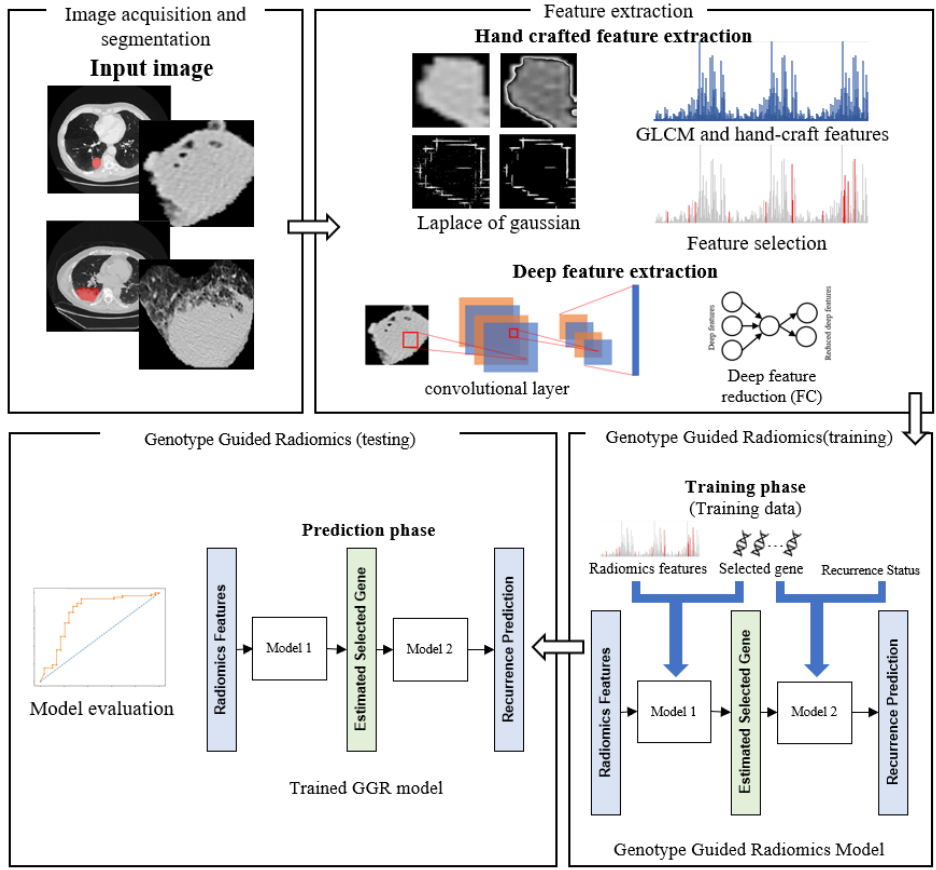

Fig. 1. Overview of genotype guided radiomics (GGR) for recurrence prediction of non-small-cell lung cancer.

to predict recurrence of NSCLC. The advantage of the proposed method is that we can significantly improve the recurrence prediction accuracy compared to direct CT-based method because we associated the gene information in the prediction models.

## 2 MATERIALS AND METHODS

The workflow of this study has been presented in Figure 1. The proposed method consists of three parts: image preprocessing; radiomics feature extraction and selection (input features); prediction model. The core is the prediction model, which is called as genotype guided radiomics (GGR) model. Unlike the traditional radiomics-based method, which uses one model to predict the recurrence from the CT images, the proposed genotype-guided radiomics (GGR) method consists of two models. The first model is used to estimate the gene expression information from CT images and the second model is used to predict the recurrence using the estimated gene expression. Both models were trained using the NSCLC public radiogenomic dataset [17], which includes both CT images and corresponding gene data. The important idea of the proposed method is that we only use CT images to predict recurrence, while the models are trained by pairs of gene data and CT images in the training phase. That means the trained GGR model can represent some relationship between the CT image and its gene expression. Therefore, we can use some gene information estimated from the CT image for recurrence prediction even though we do not have gene data in the test phase.

### 2.1 Dataset

In this research, a radiogenomics dataset of NSCLC [17], which now has an open public access in The Cancer Imaging Archive (TCIA) [18], was used. The R01 cohort from Stanford University School of Medicine and Palo Alto Veterans Affairs Healthcare System were recruited between April 7th, 2008 and September 15th, 2012. Subjects signed written consent forms according to the guidelines of institutions' IRBs. This dataset was collected from the NSCLC cohort of 162 subjects and the data of every subject can be separated into two types including image data and gene data [19, 20].

#### 2.1.1 Image data

Image data is collected from preoperative CT scans at the Stanford University Medical Center. Imaging data has a thickness of 0.625–3 mm and an X-ray tube current at 124–699 mA at 80–140 kVp. The type of CT images is in the DICOM format and consist of 162 subjects in total [18]. Some samples are screened out by some rules, including every CT image that does not have a segmented area or tumor mask, the patient with no gene data or recurrence data or died before recurrence occur due to unclear reasons.

The initial segmentation was received from axial CT-image series using the unpublished automatic segmentation algorithm [18]. All these divisions are seen by thoracic radiologists with experienced over 5 years and corrected the data as needed using ePAD software [18]. The final segmentation is revised and final approval (tumor discussion and amended as appropriate) by additional thoracic radiologists [18]. All segmentation will be stored as DICOM segmentation objects [18]. Some examples of image data and segmented tumor mask area are shown in Figure 2.

### 2.1.2 Gene data

To examine gene information, biologists must collect the tumor samples. All tumor samples were collected from drugless volunteers during surgery. After tumor removal, the surgeon cuts 3-5 mm thick pieces along the longest axis of the cut tissue, which is frozen within 30 minutes after removal. The tissue was analyzed using the expression of RNA sequences.

The RNA-sequencing (RNA-seq) data is depending on the availability and quality of existing tissues, RNA-seq analysis was performed with samples. The RNA-seq was sequenced by HiSeq 2500 (Illumina) following with the manufacturer's instructions. The 130 tissue-set samples have been sequenced in 3 batch sizes: 16, 66, and 48. The RNA-seq data was preprocessed by Centrillion Bioscience and the gene expression is estimated in Fragments per Kilobase of Transcript Per Million (FPKM) [18, 19]. 22,127 genes were provided for each patient [18]. In the dataset, there are only 130 from 162 genes of patients were present [18]. Most patients' gene data has showed unclear expression in some patients, indicated by N/A. These unclear gene expressions will be removed from our work. Finally, 5,587 genes are available in the study dataset for each patient. After data has been screened, the dataset includes the images of 88 patients which meet our experimental requirement has been remained. The dataset details are presented in Table 1.

## 2.2 Image Feature Extraction and Selection

Before feature extraction procedure, we first select the slice with the largest tumor mask area and select the adjacent above and below slices (three slices were selected in total). Here, intensity cut was performed to get the suitable radio information. We cut the intensity information outside the range of -1000 HU ~ +400 HU (Hounsfield Unit), which is covered the information that we want from CT image, including air, water and tissues [22], and normalize value in entire three slices to grey scale value with range 0 to 255 using linear transformation (Figure 5 (a)). In each CT image, the masking on the tumor area means the segmentation data attached to the data set that proceeds to multiply. Then we crop the area outside the bounding box around the masked image and resize the cropped image to 224 × 224 pixels. From this part, we get tumor area CT-image with size 224 × 224 × 3 pixels per each patient and be ready to extract the features.

TABLE 1
CLINICAL CHARACTERISTICS OF THE SCREENED SUBJECTS

|  |  | Total n = 88 patients |
|---|---|---|
| Age (year) |  | 46~85 (median = 69) |
| Gender | Male | 64 (72.72%) |
|  | Female | 24 (27.27%) |
| Smoking status | Nonsmoker | 15 (17.05%) |
|  | Current | 14 (15.91%) |
|  | Former | 59 (67.05%) |
| Cell type | ADC | 68 (77.27%) |
|  | SQC | 17 (19.30%) |
|  | Not specified | 3 (3.41%) |
| Recurrence | No | 29 (32.95%) |
|  | Yes | 59 (67.05%) |

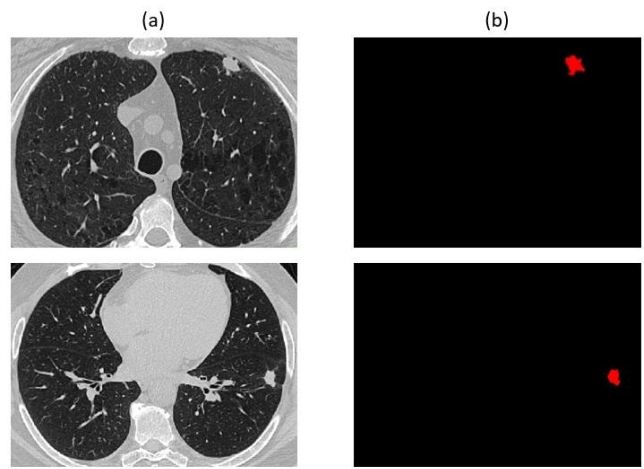

Fig. 2. Two examples of image data (a) image data and (b) tumor mask area

### 2.2.1 Handcrafted Feature Extraction

We currently have 3 slices of cropped CT-image. On each slice, 150 radiomics features of radiomics signature will be extracted using gray level co-occurrence matrix (GLCM) in 4 directions, 0, 45, 90 and 135 degrees, and histogram-based statistics [23-26]. In the calculation of each part, Laplace of Gaussian (LoG) has been performed.

The LoG is the combination between the Laplacian filter and the Gaussian filter. Technically, the Gaussian filters make images smoother to reduce noise, and Laplacian filters focus on areas that change quickly or work like edge detection. To calculate LoG, we first calculate the gaussian distribution using equation (1) [26].

$$G(x, y; \sigma) = \frac{1}{\sqrt{2\pi\sigma^2}} \exp\left(-\frac{x^2+y^2}{2\sigma^2}\right) \quad (1)$$

where x and y are the coordinates on the x and y axes, respectively, and $\sigma$ is the standard derivation or the biometrics filter parameter [22-24]. The Gaussian scale-space representation of the image $f(x, y)$ can be computed as equation (2)

$$L(x, y; \sigma) = f(x, y) * G(x, y; \sigma) \quad (2)$$

From equation (2), $L(x, y; \sigma)$ is the Gaussian scale space of the image $f(x, y)$, and * is a convolutional operator [22-24]. From both equation (1) and equation (2), we apply the Laplacian operator $\nabla^2$ to the Gaussian scale–space representation of the image. We can calculate $\nabla^2$ as equation (3).

$$\nabla^2 = \frac{\partial^2 f}{\partial x^2} + \frac{\partial^2 f}{\partial y^2} \quad (3)$$

Then, the LoG can be summarized as equation (4).

$$\nabla^2 G(x,y) = -\frac{1}{\pi\sigma^4}\left[1 - \frac{x^2+y^2}{2\sigma^2}\right]\exp^{-\frac{x^2+y^2}{2\sigma^2}} \quad (4)$$

From LoG equation (1) - (4), five different $\sigma$ filters ($\sigma = 0, 1, 1.5, 2, 2.5$; Fig. 3 (b)) were assigned to the calculation.

A four directions, which involves angles at 0°, 45°, 90°, and 135°, GLCM was calculated from each LoG filter. Radiomics features include contrast, entropy, relationships, homogeneity, and energy calculated from each degree of angle. In addition, the intensity features use first-order statistic to quantify the radiomics feature. The first order statistic includes mean, standard derivation (SD), percentiles mean (10, 25, and 50 percentiles), percentiles SD (10, 25, and 50 percentiles), kurtosis, and skewness [23]. The structure of radiomics features is shown in Fig. 4. Next, follow the mentioned procedure using the other two adjacent slices of the biggest tumor mask slice. Finally, 450 radiomics features were extracted from one patient.

After radiomics feature extraction, the extracted radiomics feature which is not related to the gene will be removed. In [3, 7–9, 23] and the conventional radiomics method, the feature selection is used to select the radiomics signature of the top association to the recurrence of the NSCLC. From the reported in [23], we found that F-test (ANOVA; Analysis of Variance) has showed highest accuracy and AUC in recurrence prediction using CT-image compared to various feature selection methods. In this work, we decide to use F-test [25-26] to select the associated radiomics features to predict the gene related to the recurrence. With F-test, we can reduce the number of radiomics features from 450 radiomics features to only 12 associated radiomics features. The selected 12 features are used to estimate the gene expression.

### 2.2.2 Deep Feature Extraction

In deep feature extraction, we applied the pretrained ResNet50 structure [13] using the ImageNet dataset [29] and eliminated the fully connected layers. ResNet50 is a convolutional neural network model that is 50 layers deep [13]. Inside the ResNet50, it consists of many residual blocks connected. In each block, it has convolution modules and a skip connection that pass the information calculated from previous block to the next block. It has been observed that in earlier layers, the learned features correspond to the lower semantic information. Without skip connection, the information will turn to abstract because of the long chain connection. In other words, non-skip connection is the cause of very small gradient in the deeper layer and have chance to become zero. From this reason, we cannot update the early layer at all. As show in Figure 4 (a), $x$ represents the input, $H(x)$ represents the output, then, the residual knowledge of the shortcut connection is $F(x) = H(x) - x$. The skip connection can reduce loss from the deep calculation, which effected by chain rule in the back propagation, because this provides the alternative path for the gradient and allow the information pass through. The information that passed the skip connection will be attached to the calculated information via addition. The structure of ResNet50 model is shown in Figure 4 (b).

The pretrained ResNet50 network using ImageNet can classify images into 1000 different object categories, such as many animals and things, using ResNet50 structure. The pretrained weight from ImageNet has been proved that it can improve the performance of the medical image analysis [30]. Removing last dense layer allow us to get several deep features that can use for genes and recurrence prediction.

The NSCLC recurrence-related output features extracted from the ResNet50 will be selected using the F-test method as the feature selection. Only around 27000 deep features were selected using F-test method. Finally, the related deep features that are ready to be the inputs of the GGR models were obtained. Some examples of convolution product from intermediate layers of the deep-features extraction (conv1 and res2a_branch2a layers) are shown in Figure 5 (c).

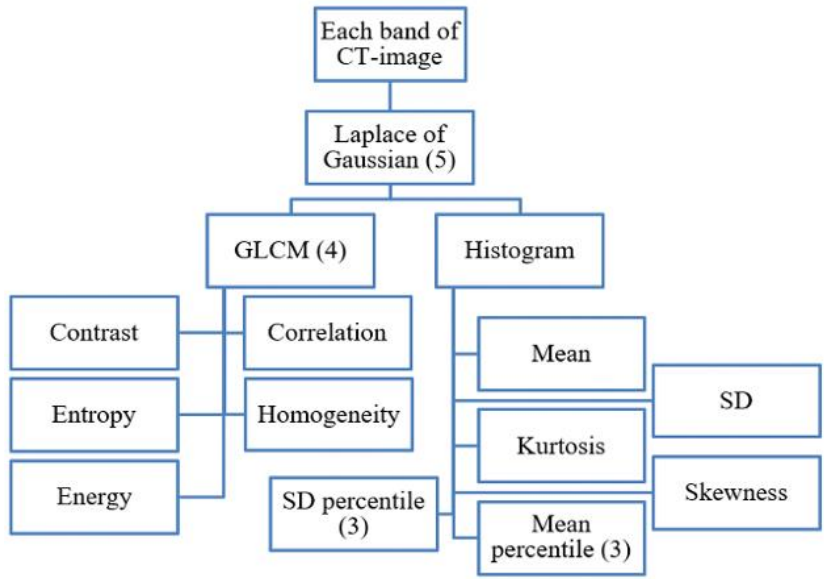

Fig. 3. The radiomics feature structure from each band of CT image. (The number in parentheses indicates the number of the features' groups.)

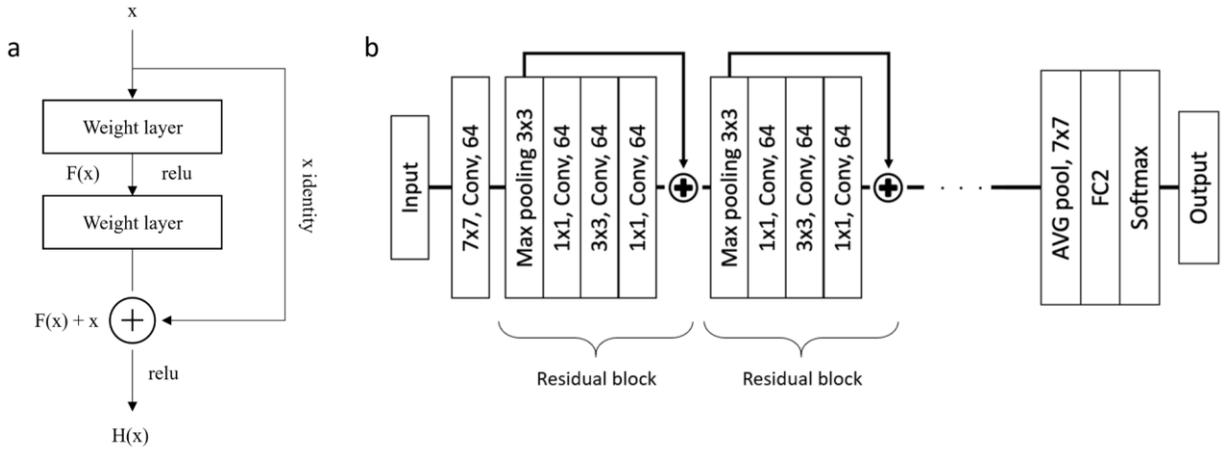

Fig 4. (a) A single residual block with the skip connection used in ResNet. (b) The ResNet structure.

## 2.3 Gene Selection

Due to the gene expression is a relatively large dataset which contain more than 20,000 gene data for each patient, some data are not related to the recurrence. Too many genes will significantly increase the computational cost and reduce the recurrence prediction results. In other words, spending time for the estimation of 20,000 genes is worthless. From this reason, we must select only associated gene that we want to estimate before training the model.

Based on the large data of genes from a patient, relevant genes need to be selected before declaring the regression model. In other word, we will not declare the regression model for non-relevant gene. By the gene selection, the genes which are not associated to recurrence of NSCLC will be removed. In conventional radiomics method [3, 7–9], the feature selection will be performed to select the radiomics signature only. In the GGR, the feature selection method is also used to select the associated genes. The methods used to select the corresponding genes in this study are including nonselected, LASSO [24, 25], F-test (ANOVA) [27, 28], CHI-2 [31] and the intersection of the three feature selection methods. Finally, we selected 74 related genes. The detailed results will be reported in Sec. 3. Hereby, we have fewer gene (only 74) to estimate.

## 2.4 Gene Estimation

We use a deep neural network (DNN) regression model to estimate gene estimation as shown in Fig.6. The inputs of the model are the selected radiomics features, which are extracted from CT images and the deep features extracted from the pretrained ResNet50 model. As we described in Sec. 2.2, we select 12 related handcrafted-based features for gene estimation from all 450 features of each patient's CT image. To balance the input of 12 handcrafted-based features and more than 27000 deep features, we use the fully connected layer to reduce the deep learning-based features to only 12 features before concatenation. The gene data is required to train the model (the mapping function between the CT image and gene data) but do not need gene data in testing phase because we can estimate the gene information from the CT image using the trained model. Note that one regression model is used to estimate one gene. As we described in Sec. 2.3, we select 74 genes, which are related to early recurrence, to reduce the computation cost and enhance the prediction accuracy. That means we trained 74 regression models for each gene. All regression models have the same structure as shown in Fig.5 but have different weights. For all the DNN regression models, we use mean square error [32] as loss function which can be calculated using equation (5).

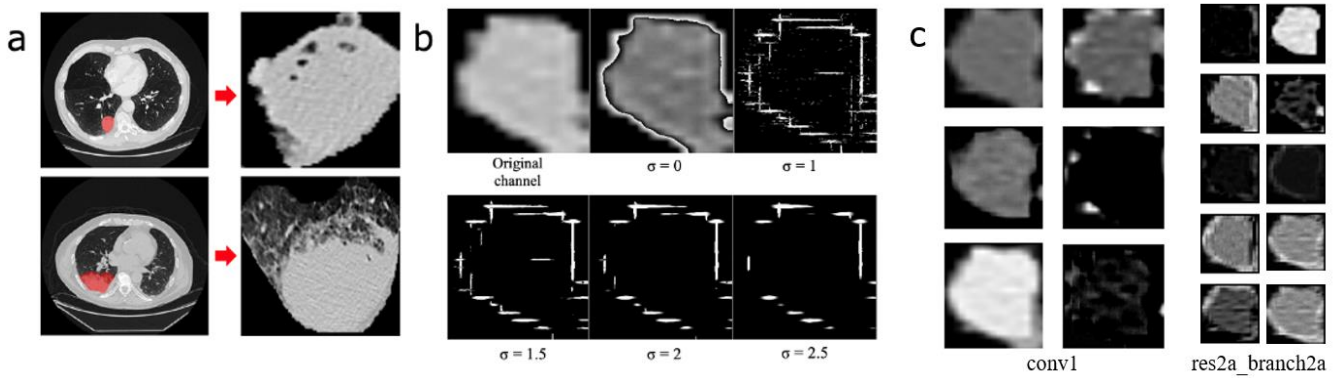

Fig. 5. (a) Full CT images with the red masking area (left) and the images after preprocessing (crop and normalization; right). (b) The result of five filters LoG on one slice. (c) The examples of convolution product from intermediate layers of deep learning-based features extraction (conv1 and res2a_branch2a layers).

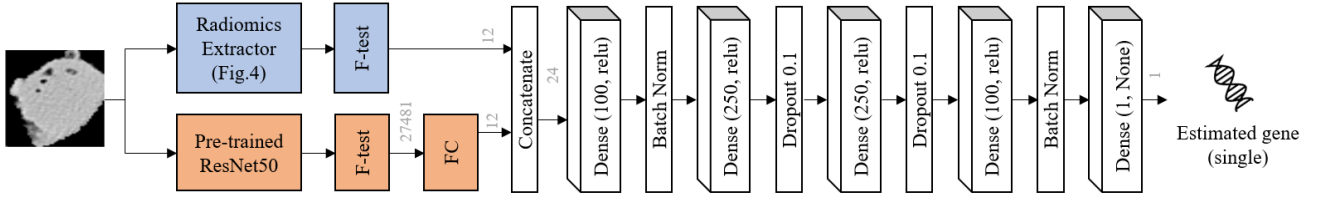

Fig. 6. The DNN regression model's structure used to estimate single gene from the extracted and selected features

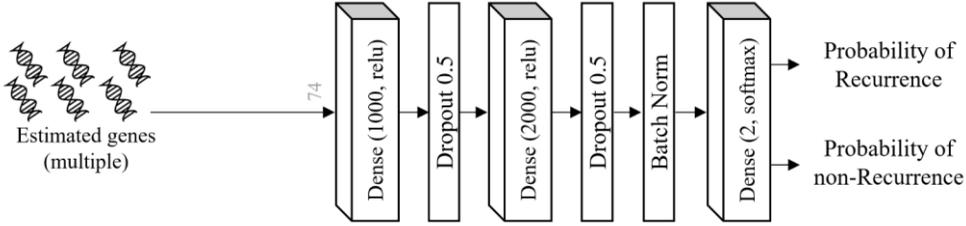

Fig. 7. The GGR's structure used to predict the recurrence using estimated genes. The grey numbers show the feature's number that pass the process. The grey numbers show the feature's number that pass the process.

$$\mathcal{L}_{mse} = \frac{1}{m}\sum_{i=1}^{m}(g_i - \hat{g}_i)^2 \qquad (5)$$

where $\mathcal{L}_{mse}$ denoted mean squared error loss, m denoted the number of data points, g denoted the actual gene expression value and $\hat{g}$ denoted the gene prediction. The learning rate was set to very small value of 5e-6. The decay and momentum were set to 1e-6 and 0.9, respectively. The DNN regression model's structure used to estimate a gene is shown in Figure 6.

### 2.4 The Recurrence Prediction

The recurrence prediction was done as a task of two-class classification (i.e., recurrence or non-recurrence). We proposed second model (model 2) to predict the recurrence of NSCLC using estimated genes as shown in Figure 7 We use stochastic gradient descent (SGD) [33] as optimizer. The learning rate, decay and momentum were set to 0.05, 1e-6 and 0.9, respectively. The loss function used in GGR is binary cross entropy loss [34] which can be calculated by using equation (6).

$$\mathcal{L}_{ce} = -\frac{1}{m}\sum_{i=1}^{m}[r_i \ln \hat{r}_i + (1 - r_i)\ln(1 - \hat{r}_i)] \qquad (6)$$

where $\mathcal{L}_{ce}$ denoted binary cross entropy loss, m denoted number of the data points, $r$ denoted actual recurrence status and $\hat{r}$ denoted recurrence status prediction.

## 3 EXPERIMENTS

Every experiment is performed based on ten-fold cross validation for five times to find the average performance. On each fold, CT-images and gene data of 79–80 patients are used as training sets and the CT-image of 7-8 patients will be set as the validation set. In addition, gene data is not required for real use. There is also an area calculation under the characteristic curve of the receiver operating characteristics (AUC) [35] to assess the efficiency. We performed these process on a personal computer driven by CPU Intel® core™ i7-8700k @3.20-4.60GHz, 48 GB of random-access memory (RAM) and RTX 2060 graphic accelerator. We used Keras-GPU library version 2.2.4 on python 3.6 to perform these experiment's actions. The related work models are compared to the proposed GGR methods using the same control data set.

### 3.1 Gene Selection Results

This part is used to choose the best gene set, most relevant to the recurrence of NSCLC. In four feature selection methods, including nonselected, LASSO, F-test, CHI-2, and the intersection of the three, the gene features which have LASSO's zero coefficient were removed [22, 23]. The P-values from F-test and CHI-2 were set the threshold at P-value < 0.02 [25-26, 28]. The intersection of the three uses the results from the three methods to make the intersection of all selected genes. Each output gene set will be tested by classifying the recurrence. The gene set from the best feature selection method will be chosen to predict the recurrence in model 2. The results of gene selection performance are presented in Table 2. Since the method of intersection of the three achieves the best performfoance, we used 74 genes selected by F-test for our studies. The 74 selected genes list has shown in Table 3.

TABLE 2
PERFORMANCE OF THE RADIOMICS METHOD USING GENE DATA.

| Feature selection method | Selected genes | Accuracy |
|---|---|---|
| Nonselected | 5587 | 0.8141 |
| LASSO | 1123 | 0.8297 |
| F-test | 131 | 0.8689 |
| CHI-2 | 2325 | 0.8339 |
| Intersection of LASSO, F-test, and CHI-2 (P-value < 0.02) | **74** | **0.8806** |

TABLE 3
THE 74 GENES SELECTED BY THE INTERSECTION OF THE THREE METHODS (P-VALUE < 0.02).

| Gene list | | | | | | | | | |
|---|---|---|---|---|---|---|---|---|---|
| ABCC9 | ANKLE2 | ANKRD13D | ANKZF1 | AP1G2 | ARRDC2 | ATAD2 | ATG4B | ATIC | BARD1 |
| BCL6 | BMS1 | BRAP | BTG2 | CBX5 | CD2AP | CEBPZ | CIRBP | CREBL2 | CRK |
| CYB5A | CYBRD1 | DCAF13 | DCLRE1C | DDX42 | DDX51 | DENR | DHCR24 | DNAJB6 | DNPEP |
| DPYSL2 | DROSHA | EMP2 | ENY2 | ERC1 | ERCC6 | ETNK1 | FAM122B | FAM20B | FANCL |
| FOXK2 | FRMD4B | GAK | GCN1L1 | GGA3 | GLUD1 | GMDS | GTPBP2 | GTPBP4 | HIPK3 |
| HIST1H1D | HIST1H4B | HIST1H4C | IL6ST | KIAA0368 | LAMB2 | LOC100133091 | LPAR6 | LRPPRC | MAPK14 |
| MAT2A | MBOAT2 | MCM4 | MCM7 | MYO10 | NAA25 | NOP56 | PTGES3 | RBMS3 | RICTOR |
| SELENBP1 | SNORA37 | SPG7 | TXNIP | | | | | | |

## 3.2 GENE ESTIMATION RESULTS

This section shows the comparison between the estimated genes derived from the model and the real gene values. Due to the complexity of the data, a comparison graph between the estimated genes using the hybrid of handcrafted based and deep features, and the actual values of one sample has shown in Figure 8. From Figure 8, x-axis show 74 genes of one sample patient, which are selected by the combination of the three gene selection methods, and y-axis show the actual gene expression (the blue line) value, the estimated gene expression from the combination of handcrafted features and the deep features (the orange line; average MSE = 29293.07). The graph show that the majority of the estimated genes show satisfying estimation performance.

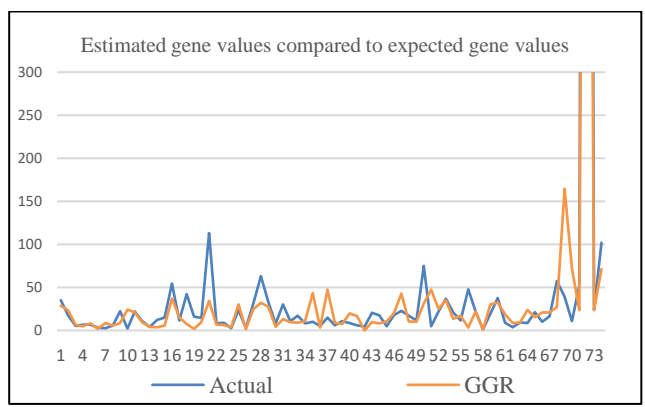

Fig. 8. Comparison between estimated gene values and actual gene values from one patient as an example.

## 3.3 RECURRENCE PREDICTION RESULTS

This experimental part is used to investigate how the proposed GGR models can appropriately work compared with the other related works in terms of accuracy and AUC (Table 4). The proposed methods are indicated in bold. Furthermore, Figure 9 show the comparison between the average ROC curves of 10-fold cross validation to the other recurrence prediction methods. As shown in the Table IV the proposed GGR method outperforms the traditional radiomics-based method [8] and deep learning-based methods [12-16], since the proposed models can learn more useful features by using the genotype-guidance in the training phase.

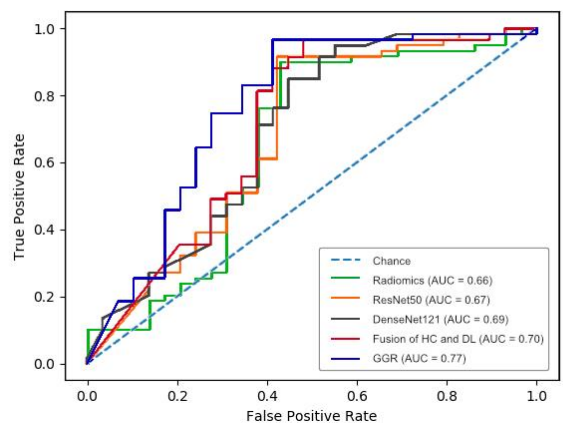

Fig. 9. Average ROC of recurrence prediction using conventional radiomics method, DL-based radiomics (ResNet50), DL-based (DenseNet121), fusion of handcrafted and deep features radiomics, and proposed GGR.

## 4. ABLATION STUDY

In the GGR, we use the fusion of the handcrafted and deep features to the recurrence prediction. To evaluate the effectiveness of each feature type, we study ablation experiments with the same control dataset. The gene estimation of the standalone handcrafted GGR and the deep GGR are measured in mean square error (MSE). The MSE of the standalone handcrafted GGR and deep GGR are 48508.45 and 50614.05, respectively. The accuracy of the standalone handcrafted features GGR and deep features GGR are 81.56% (AUC = 0.73, specificity is 0.57 and sensitivity is 0.93) and 76.25% (AUC = 0.64, specificity is 0.38 and sensitivity is 0.95), respectively. Both standalone handcrafted GGR and deep GGR cannot reach the fusion of handcrafted and deep GGR performance in both gene estimation and recurrence prediction models. This ablation experiment show that the fusion of both feature types is effective and essential.

TABLE 4
THE PERFORMANCE OF GGR COMPARED WITH THE STATE-OF-THE-ART METHODS. THE PROPOSED METHODS ARE INDICATED IN BOLD.

| Method | AUC | Specificity | Sensitivity | Accuracy |
|---|---|---|---|---|
| Using CT radiomics signatures [8] | 0.6567 | 0.56 | 0.90 | 78.61% |
| DL-based Radiomics (ResNet50) [14] | 0.6714 | 0.59 | 0.89 | 79.09% |
| Fusion of HC and DL-based [12] | 0.7078 | 0.51 | 0.97 | 82.08% |
| **Genotype Guided Radiomics (proposed method)** | **0.7667** | **0.59** | **0.95** | **83.28%** |

## 5. DISCUSSION AND CONCLUSION

In this paper, we studied the recurrence of NSCLC prediction using CT-image analysis to help the doctor and patient to prepare for the risks that may occur. Since the traditional radiomics-based methods or recently proposed deep learning-based methods used only CT information, the prediction performance was limited. From the experiments, we can achieve 78.61% accuracy (AUC = 0.6567) for handcrafted based radiomics, 79.09% accuracy (AUC = 0.6714) for deep learning-based (ResNet50). The prediction accuracy and AUC were significantly improved to 83.28% (AUC = 0.7667) by the proposed GGR. Since the proposed prediction methods used genotype information to guide the model training, the prediction models can learn more useful information for early recurrence prediction though the proposed method also only uses CT information for prediction in the test phase. We also performed experiments with real gene expression for early recurrence prediction. From our experiments, we can achieve 91.83% accuracy (AUC = 0.9210) for gene-expression analysis and 93.19% accuracy (AUC = 0.9268) for the fusing between gene-expression and CT radiomics signature, which are much higher than the proposed method. As our future work, we are going to improve the models and increase training dataset to improve the predication performance and make them closer to the genomics-based methods but with only CT images.

## ACKNOWLEDGMENT

This work was supported in part by the Grant-in Aid for Scientific Research from the Japanese Ministry for Education, Science, Culture and Sports (MEXT) under the Grant No. 20KK0234, No.18H03267 and No. 20K21821.

**Panyanat Aonpong** was born in Suphanburi Province, Thailand, in 1993. He

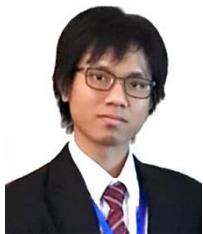

received a Bachelor of Engineering degree, Computer Engineering, from King Mongkut's Institute of Technology Ladkrabang, Bangkok, Thailand, in 2015. Master of Engineering degree, Information Communication Technology for Embedded Systems from Kasetsart University, Bangkok, Thailand, in 2017. Now, he is studying at the doctoral degree in Department of Information Science and Engineering, Ritsumeikan University, Shiga, Japan, since 2018. He is interested in computer image processing. He has experience working in satellite imagery when he was working in Geoinformatics Information and Space Technology Development Agency (GISTDA), the Thai government organization which works directly to the satellites and images.

**Yutaro Iwamoto** received the B.E. and M.E., and D.E. degree from

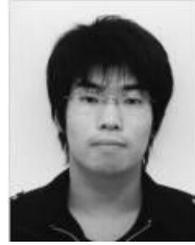

Ritsumeikan University, Kusatsu, Japan in 2011 and 2013, and 2017, respectively. He is currently an Assistant Professor at Ritsumeikan University, Kusatsu, Japan. His current research interests include image restoration, segmentation, classification of medical imaging and deep learning

**Xian-Hua Han** received a B.E. degree from ChongQing University,

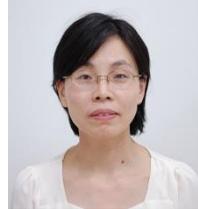

ChongQing, China, a M.E. degree from ShanDONG University, JiNan, China, a D.E. degree in 2005, from the University of Ryukyus, Okinawa, Japan. From April 2007 to March 2013, she was a postdoctoral fellow and an associate professor with College of Information Science and Engineering, Ritsumeikan University, Shiga, Japan. From April 2016 to March 2017, she was a senior researcher with the Artificial Intelligence Research Center, National Institute of Advanced Industrial Science and Technology, Japan. She is currently an associate Professor with the Artificial Intelligence Research Center, Yamaguchi University, Japan. Her current research interests include image processing and analysis, pattern recognition, machine learning, computer vision and hyperspectral image analysis. She is a member of the IEEE, IEICE.

**Lan-fen Lin** was born in Pingyang, China in 1969. She received her B.S. and

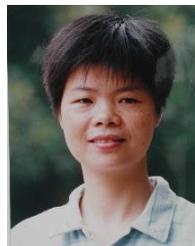

Ph.D. degrees in Aircraft Manufacture Engineering from Northwestern Polytechnical University in 1990, and 1995 respectively. She was post doctor in the college of Computer Science and Technology of ZJU form Jan. 1996 to Dec. 1997. Now she is a full professor at the college of Computer Science and Technology, Zhejiang University, China, vice director of Artificial Intelligence Institute in ZJU., and member of Advisory Expert Group for manufacturing industry informatization in Zhejiang Province. Her research interests include Big data Analysis, Data Mining, Knowledge Management, etc. She is currently leading a research group to exploring AI technologies for medical imaging. She is the leader of more than 40 projects, include the subproject of Major State Basic Research Program (973) of china, and the National High Technology Research and Development Program of China. Lan-fen Lin has published more than 140 research papers in the journals or international conference proceedings.

**Yen-Wei Chen** was born in Hangzhou, China in 1962. He received his B.E.

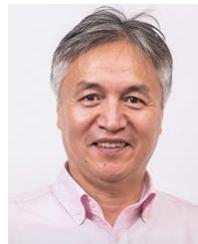

degree in 1985 from Kobe Univ., Kobe, Japan. He received his M.E. degree in 1987, and his D.E. degree in 1990, both from Osaka Univ., Osaka, Japan. He was a research fellow at the Institute of Laser Technology, Osaka, from 1991 to 1994. From Oct. 1994 to Mar. 2004, he was an associate Professor and a professor in the department of Electrical and Electronic Engineering, Univ. of the Ryukyus, Okinawa, Japan. He is currently a professor at the college of Information Science and Engineering, Ritsumeikan University, Japan. He is also a visiting professor at the college of Computer Science and Technology, Zhejiang University, and Zhejiang Lab, Hangzhou, China. His research interests include medical image analysis, pattern recognition. Yen-Wei Chen has published more than 300 research papers. He has received many distinguished awards. He is the Principal Investigator of several projects in biomedical engineering and image analysis, funded by Japanese Government.